\documentclass[conference]{IEEEtran}
\IEEEoverridecommandlockouts
\usepackage{cite}
\usepackage{amsmath,amssymb,amsfonts}
\usepackage{algorithm}
\usepackage{algpseudocode}
\usepackage{graphicx}
\usepackage{makecell}
\usepackage{textcomp}
\usepackage{comment}
\usepackage{xcolor}
\usepackage{lipsum}  
\usepackage[mathscr]{euscript}
\usepackage{dblfloatfix}
\usepackage{cuted}
\usepackage{multirow}
\def\BibTeX{{\rm B\kern-.05em{\sc i\kern-.025em b}\kern-.08em
    T\kern-.1667em\lower.7ex\hbox{E}\kern-.125emX}}
\begin{document}
\algrenewcommand\algorithmicindent{0.8em}
\title{Dynamic Continual Learning: Harnessing Parameter Uncertainty for Improved Network Adaptation\\
\thanks{This work was supported by National Science Foundation (NSF) Awards NSF ECCS 1903466 and NSF OAC 2008690.
}
}

\author{\IEEEauthorblockN{Christopher F. Angelini}
\IEEEauthorblockA{\textit{Dept. of Electrical and Computer Engineering} \\
\textit{Rowan University}\\
Glassboro, New Jersey, USA \\
angelinic0@rowan.edu}
\and
\IEEEauthorblockN{Nidhal C. Bouaynaya}
\IEEEauthorblockA{\textit{Dept. of Electrical and Computer Engineering} \\
\textit{Rowan University}\\
Glassboro, New Jersey, USA \\
bouaynaya@rowan.edu}
}
\maketitle

\begin{abstract}
When fine-tuning Deep Neural Networks (DNNs) to new data, DNNs are prone to overwriting network parameters required for task-specific functionality on previously learned tasks, resulting in a loss of performance on those tasks. We propose using parameter-based uncertainty to determine which parameters are relevant to a network's learned function and regularize training to prevent change in these important parameters. We approach this regularization in two ways: (1), we constrain critical parameters from significant changes by associating more critical parameters with lower learning rates, thereby limiting alterations in those parameters; (2), important parameters are restricted from change by imposing a higher regularization weighting, causing parameters to revert to their states prior to the learning of subsequent tasks. We leverage a Bayesian Moment Propagation framework which learns network parameters concurrently with their associated uncertainties while allowing each parameter to contribute uncertainty to the network's predictive distribution, avoiding the pitfalls of existing sampling-based methods. The proposed approach is evaluated for common sequential benchmark datasets and compared to existing published approaches from the Continual Learning community. Ultimately, we show improved Continual Learning performance for Average Test Accuracy and Backward Transfer metrics compared to sampling-based methods and other non-uncertainty-based approaches. 
\end{abstract}


\section{Introduction}
The term ``Narrow AI'' is gaining traction for describing Artificial Intelligence (AI) and Machine Learning (ML) systems that cannot adapt to information after deployment. In standard training schemes, Deep Neural Networks (DNNs) assume the collection of observations a network is trained on will accurately describe the environment in which it is deployed. In the real world, DNNs are regularly subject to out-of-distribution (OOD) data, various types of noise, shifting distributions of conceptual objectives, and may require adaptation to new data after the initial training period \cite{ditzler_learning_2015}. These characteristics of real-world data undermine assumptions of data consistency made during training and require DNNs to have the ability to adapt and expand upon previously learned data representations. 

Unlike biological learning systems, which can adapt and consolidate learned information at will, standard machine learning systems restrict network performance to the most recently trained task \cite{kirkpatrick_overcoming_2017, parisi_continual_2019}. Fine-tuning networks to new information generally results in a partial or complete loss of performance on previously trained tasks, known as \textit{Catastrophic Forgetting} or \textit{Catastrophic Interference} \cite{mccloskey_catastrophic_1989,french_catastrophic_1999}. This phenomenon arises from overwriting and replacing network parameters during the training process for a new task. As a result, a sub-discipline of machine learning called \textit{Continual Learning} (CL), or \textit{Lifelong Learning}, has emerged, focusing on mitigating Catastrophic Interference in DNNs over a sequence of tasks.

Current approaches to CL can be categorized into four main categories: Regularization, Dynamic Architectures, Rehearsal, and Dual Memory Systems \cite{parisi_continual_2019}. Regularization-based approaches restrict parameter change when training subsequent tasks attempting to preserve previous network representations \cite{kirkpatrick_overcoming_2017,zenke_continual_2017,aljundi_memory_2018,serra_overcoming_2018}. Dynamic Architecture approaches add neural resources to expand on previous representations \cite{yoon_lifelong_2018,rusu_progressive_2022}. Rehearsal approaches preserve encompassing data samples that describe most information learned from previous tasks, supplementing training new tasks with these selected data samples \cite{rebuffi_icarl_2017,hurtado_memory_2023, belouadah_il2m_2019}. Dual Memory Systems consolidate information from short-term, temporary to long-term, permanent memory to enable rapid adaptation while preserving previously encountered information \cite{kemker_fearnet_2018,chaudhry_efficient_2019,lopez-paz_gradient_2022,shin_continual_2017}. 

Methods described in this paper will focus on regularization-based approaches to avoid undesirable characteristics of other approaches, such as storing previous data, adding neural resources for new tasks, or complex network and optimization structures. At the limit, allowing for such characteristics would create an ever-growing storage or computational resource requirement unsuitable for many real-world applications. Regularization-based approaches focus on constraining the optimization process when learning a new task to avoid significant changes to parameters learned in previous tasks. Most regularization approaches rely on identifying important parameters within the network and adjusting regularization according to the relevance of each parameter to a previous task. This contrasts with applying a uniform factor across all network parameters.

This paper details two new regularization approaches to CL by managing changes in network parameters with learned parameter uncertainty inherent to Bayesian Deep Learning networks. The learned statistical uncertainty derived from the approximated posterior distribution serves as the basis for determining the importance of each network parameter in the context of a previously learned task. Contrary to existing Bayesian approaches that rely on Monte Carlo sampling to estimate network uncertainty, we developed a network-agnostic framework, called Moment Propagation (MP) \cite{dera_premium-cnn_2021}, to learn the values of network parameters and their associated uncertainty in an online manner, without resorting to computationally expensive sampling techniques. In contrast to previous approaches to deep Variational Inference (VI) \cite{kingma_auto-encoding_2022,blundell_weight_2015}, MP networks approximates the network's predictive distribution by propagating the first two moments, mean and covariance, through the entire network. The first-order Taylor Series approximation is used to enable propagation of covariance through non-linearities in the network i.e. non-linear activation functions such as ReLU. This approach facilitates the direct maximization of the Evidence Lower Bound Objective without sampling the network likelihood \cite{dera_premium-cnn_2021}. 

We leverage parameter uncertainty learned from a MP network to govern our CL process in two ways: Learning Rate Adaptation and Per-parameter Bayesian-based regularization. Our contributions are as follows:
\begin{itemize}
    \item We demonstrate the MP framework's ability to autonomously identify important network parameters relevant to the trained network function. 
    \item We leverage the learned parameter uncertainties to regularize the training of new tasks, preventing the occurrence of catastrophic forgetting of previous tasks.
    \item We demonstrate catastrophic forgetting mitigation in the task incremental learning setting with multiple sequential benchmark datasets.
    \item We compare the results of our approach with state-of-the-art approaches and baseline metrics for each network.
\end{itemize}

\section{Moment Propagation Framework}
MP leverages principles from VI to approximate the variational posterior of a network by imposing an approximating distribution, also known as variational distribution, $q_\theta(\mathbf{\Omega})$, over the network parameters. The variational distribution is then optimized by  minimizing the Kullback-Leibler (KL) divergence between the variational distribution and the true posterior distribution, $p(\mathbf{\Omega}|\mathcal{D})$, where $\mathcal{D}=\{\mathbf{X}^{(i)},\mathbf{y}^{(i)}\}{\substack{n\quad\\i=1}}$, represents the training data with $i^{\text{th}}$ input $\mathbf{X}^{(i)}$ and corresponding label $\mathbf{y}^{(i)}$. However, given that minimizing the KL divergence between the variational distribution and the true posterior is intractable due to the log evidence, an equivalent form must be considered. The log evidence can be isolated as it does not depend on the variational distribution nor will it affect the optimization. A tractable and equivalent objective called the Evidence Lower Bound (ELBO) can be maximized in place of the original KL divergence between the variational distribution and the true posterior distribution, shown in Equation (\ref{eq:ELBO_der}).
\begin{equation}\label{eq:ELBO_der}
\begin{split}
    \text{ELBO} &=\mathop{\mathbb{E}_{q_\theta(\mathbf{\Omega})}}[\ln{p(\mathcal{D}|
    \mathbf{\Omega})}] - \text{KL}_{q_\theta(\mathbf{\Omega})}[q_{\theta}(\mathbf{\Omega})||p(\mathbf{\Omega})]    
\end{split}
\end{equation}

When maximizing the ELBO it's components, the network log-likelihood, $\mathop{\mathbb{E}_{q_\theta(\mathbf{\Omega})}}[\ln{p(\mathcal{D}|\mathbf{\Omega})}]$, is maximized and the KL divergence between the network prior and the variational distribution, $\text{KL}_{q_\theta(\mathbf{\Omega})}[q_{\theta}(\mathbf{\Omega})||p(\mathbf{\Omega})]$, is minimized. Existing approaches to VI for DNNs \cite{ blundell_weight_2015} rely on approximating the network's log-likelihood by creating variations in the predictive distribution through Monte Carlo sampling of the approximated variational posterior. A distribution is then fit over these predictions to approximate the predictive distribution, allowing for the estimation of the network log-likelihood and the resulting gradient update of the network parameters. However, this process requires performing inference for each requested sample, increasing the time and computation requirements as the number of samples increase producing a trade-off between network efficacy and efficiency. 

This trade-off can be circumvented by allowing parameters to contribute their learned uncertainty to features as they are transmitted through the network, thereby accumulating uncertainty in the network's predictive distribution. We introduced a framework that propagates the first two moments of the predictive distribution through the non-linear layers of the network using a first-order Taylor Series approximation for variance, called \textit{Moment Propagation} (MP) \cite{dera_premium-cnn_2021}. This method involves learning the mean and variance of network parameters in an online fashion while enabling these parameters to influence the predictive mean and covariance. MP facilitates the propagation of an analytical expression for both the mean and covariance moments, allowing the predictive distribution to be determined without the need for sampling. Consequently, the stochasticity typically associated with the sampling process is removed from the predictive distribution estimation permitting the uncertainty in the predictive distribution to be solely based on uncertainty contributed by the network parameters. The propagation of uncertainty through this method provides an effective, deterministic measure for both mean and covariance, yielding a consistent and repeatable assessment of predictive uncertainty that is directly differentiable. Overall, MP enhances estimates of both posterior and predictive distributions with unbiased, directly differentiable evaluations of the network log-likelihood.

In the following sections, MP is derived for a convolutional neural network with $L$ layers. To streamline notation, the reference to the layer $l$ is excluded from the representation for the $l^{\text{th}}$ layer. The derivations for various network layers are presented while assuming the following:
\begin{itemize}
    \item Without loss of generality, the input feature to the network at layer $l=0$, convolutional or linear, is treated as deterministic.
    \item The $j^{th}$ network parameter $w_j$ follows a Normal distribution $w_j\sim\mathcal{N}(\mu_{w_j},\sigma_{w_j}^2)$. 
    \item The network parameters are independent of each other and the input. 
\end{itemize}

\subsubsection{Propagation through the $l^{\text{th}}$ 2D convolutional layer}
Let $\boldsymbol{\mathscr{G}} = \boldsymbol{\mathscr{W}} * \boldsymbol{\mathscr{X}} + \boldsymbol{b}$, where, $*$ denotes the convolution operation. For the $l^{\text{th}}$ convolutional layer, let $\boldsymbol{\mathscr{X}} \in \mathbb{R}^{n_1 \times n_2 \times ch}$ as a random tensor denoting the layer input, $\boldsymbol{\mathscr{W}} \in \mathbb{R}^{k_1 \times k_2 \times ch \times f}$ as a random tensor denoting the layer weights, $\boldsymbol{b} \in \mathbb{R}^{f}$ as a random vector of denoting the layer biases, and $\boldsymbol{\mathscr{G}} \in \mathbb{R}^{out_1 \times out_2 \times f}$ as random tensor denoting the propagated feature.

We consider the vectorized form of each filter within weight random tensor $\boldsymbol{\mathscr{W}}$ reforming the random tensor to matrix $\boldsymbol{W} \in \mathbb{R}^{k \cdot k \cdot ch \times f}$.
Let $\boldsymbol{W}=[\boldsymbol{w}_1, \cdots, \boldsymbol{w}_f]$, where $\boldsymbol{w}_i$ is the $i^{\text{th}}$ column of $\boldsymbol{W}$, representing the $i^{\text{th}}$ vectorized filter with the mean and covariance $\boldsymbol{\mu}_{\boldsymbol{w}_i} \in \mathbb{R}^{k \cdot k \cdot ch }$ and $\boldsymbol{\Sigma}_{\boldsymbol{w}_{i}} \in \mathbb{R}^{k \cdot k \cdot ch \times k \cdot k \cdot ch}$, respectively. 
Similarly, we consider the vectorized form of the image patches under each filter within the input random tensor $\boldsymbol{\mathscr{X}}$ reforming the random tensor to matrix $\boldsymbol{X} \in \mathbb{R}^{k \cdot k \cdot ch \times out_1 \cdot out_2}$.
Let $\boldsymbol{X}=[\boldsymbol{x}_1, \cdots, \boldsymbol{x}_{[out_1 \cdot out_2]}]$,where $\boldsymbol{x}_o$ is the $o^{\text{th}}$ column of $\boldsymbol{X}$, representing the $o^{\text{th}}$ vectorized image patch with the mean and covariance $\boldsymbol{\mu}_{\boldsymbol{x}_o}\in \mathbb{R}^{k \cdot k \cdot ch}$ and $\boldsymbol{\Sigma}_{\boldsymbol{x}_{o}}\in \mathbb{R}^{k \cdot k \cdot ch \times k \cdot k \cdot ch} $, respectively. 
It follows that the resulting random element $g_{o,i}$ with mean and covariance elements, $\mu_{g_{o,i}}$ and $\sigma_{g_{o,i}}$, contained within the resulting random matrix $\boldsymbol{G}$ can be derived for each filter $i=1 \cdots f$ and output feature pixel $o=1 \cdots (out_1 \cdot out_2)$ with the matrix-vector multiplication as shown in Equation (\ref{eq:convprop}). 
\begin{equation}\label{eq:convprop}
    \begin{split}
    \mu_{g_{o,i}} &= \boldsymbol{\mu}_{\boldsymbol{w}_i}^T \boldsymbol{\mu}_{\boldsymbol{x}_o} + \mu_{b_{i}} \\
    \sigma^2_{g_{o,i}} &=  \text{tr}\!\left(\boldsymbol{\Sigma}_{\boldsymbol{x}_o} \boldsymbol{\Sigma}_{\boldsymbol{w}_{i}}\right)\! +\!
    {{\boldsymbol{\mu}}^{T}_{\boldsymbol{x}_o}}  {\boldsymbol{\Sigma}_{\boldsymbol{w}_{i}}} {\boldsymbol{\mu}}_{\boldsymbol{x}_o}\!\! + \boldsymbol{\mu}_{\boldsymbol{w}_i}^T  \boldsymbol{\Sigma}_{\boldsymbol{x}_o} \boldsymbol{\mu}_{\boldsymbol{w}_i}\!  + {\sigma^2_{b_{i}}} \\
    \end{split}
\end{equation}
\subsubsection{Propagation through the $k^{\text{th}}$ linear layer}
Let $\boldsymbol{z} = {\boldsymbol{W}}^T \boldsymbol{x} + \boldsymbol{b}$, where, for the $k^{\text{th}}$ layer, $\boldsymbol{W} \in \mathbb{R}^{n \times m}$ is a random matrix of weights, $\boldsymbol{b} \in \mathbb{R}^{m}$ is a random vector of biases, and $\boldsymbol{z} \in \mathbb{R}^{m}$ is the resulting random vector. Let $\boldsymbol{W}=[\boldsymbol{w}_1, \cdots, \boldsymbol{w}_m]$, where $\boldsymbol{w}_i$ is the $i^{\text{th}}$ column of $\boldsymbol{W}$, with the mean and covariance of $\boldsymbol{w}_i$ represented as $\boldsymbol{\mu}_{w_i} \in \mathbb{R}^{n}$ and $\boldsymbol{\Sigma}_{w_{i}} \in \mathbb{R}^{n \times n}$. The random input vector is represented by $\boldsymbol{x} \in \mathbb{R}^{n}$ with mean and variance value vectors $\boldsymbol{\mu}_{\boldsymbol{x}} \in \mathbb{R}^{n}$ and $\boldsymbol{\Sigma}_{\boldsymbol{x}} \in \mathbb{R}^{n \times n}$, respectively. It follows that the mean and variance elements, $\mu_{z_i}$ and $\sigma^{2}_{z_i}$, contained within the resulting random vector $\boldsymbol{z}$ can be derived for elements $i=1 \cdots m$, with the matrix-vector multiplication as shown in Equation (\ref{eq:prop}). 
\begin{equation}\label{eq:prop}
    \begin{split}
    {\mu_{z_{i}}} &= \boldsymbol{\mu}_{w_i}^T \boldsymbol{\mu}_{x} + \mu_{b_{i}} \\
    {\sigma^{2}_{z_{i}}} &=  \text{tr}\left(\boldsymbol{\Sigma}_x \boldsymbol{\Sigma}_{w_{i}}\right) +
    {{\boldsymbol{\mu}}^{T}_{x}}  {\boldsymbol{\Sigma}_{w_{i}}} {\boldsymbol{\mu}}_{x} +
    \boldsymbol{\mu}_{w_i}^T  \boldsymbol{\Sigma}_x \boldsymbol{\mu}_{w_i}  + {\sigma^2_{b_{i}}} \\
    \end{split}
\end{equation}
\subsubsection{Propagation through $p^{\text{th}}$ Batch Normalization}
Let $\boldsymbol{F} = \boldsymbol{\gamma} \odot \boldsymbol{\hat{X}} + \boldsymbol{\beta}$, where $\boldsymbol{\hat{X}}$ is the normalized input $\boldsymbol{X}$ according to Equation \ref{eq:batchNormNorm}. Let input $\boldsymbol{X}=[\boldsymbol{x}_1, \cdots, \boldsymbol{x}_B]$ and $\boldsymbol{x}_b$ is the $b^{\text{th}}$ batched random vector of $\boldsymbol{X}$. Each random input vector is represented by $\boldsymbol{x_b} \in \mathbb{R}^{n}$ with mean and variance value vectors $\boldsymbol{\mu}_{\boldsymbol{x}_b} \in \mathbb{R}^{n}$ and $\boldsymbol{\Sigma}_{\boldsymbol{x}_b} \in \mathbb{R}^{n \times n}$, respectively. 
Each input vector is normalized according to Equation \ref{eq:batchNormNorm}, where $\boldsymbol{\mu_{N} \in \mathbb{R}^{n}}$ and $\boldsymbol{\sigma^2_{N} \in \mathbb{R}^{n}}$ represent the feature-wise (channel-wise in the case of 2D batch normalization) mean and variance over all $\boldsymbol{\mu}_{\boldsymbol{x}_b}$. The normalized moment vectors of mean and variance are represented by $\boldsymbol{\mu}_{\hat{x}_b}$ and $\boldsymbol{\Sigma}_{\hat{x}_b}$, correspond to $\boldsymbol{x}_b$ in $\boldsymbol{\hat{X}} = [\boldsymbol{x}_1, \cdots, \boldsymbol{x}_B]$. A small value, $\epsilon$ is added to the denominators for numerical stability.

\begin{equation}\label{eq:batchNormNorm}
    \boldsymbol{\mu}_{\hat{x}_b} = \frac{\boldsymbol{\mu_{x_b}} - \boldsymbol{\mu_{N}}}{\sqrt{\boldsymbol{\sigma^2_{N}} + \epsilon}} \\ \quad\quad\quad\quad
    \boldsymbol{\Sigma}_{\hat{x}_b} = \frac{\boldsymbol{\Sigma}_{x_b}}{\boldsymbol{\sigma^2_{N}} + \epsilon} \\
\end{equation}

\subsubsection{Propagation through a non-linear activation}
Let $\boldsymbol{\mathscr{Z}}=\Psi(\boldsymbol{\mathscr{G}})$ represent some non-linear activation function (e.g. ReLU, Hyperbolic Tangent, Softmax) of a random vector input $\boldsymbol{\mathscr{G}} \in \mathbb{R}^{m \times m \times ch}$ with mean $\boldsymbol{\mu}_z$ and covariance $\boldsymbol{\Sigma}_z$. The mean and covariance for the resulting random vector $\boldsymbol{g}$ can be approximated using the first-order Taylor series approximation \cite{dera_premium-cnn_2021} in Equation (\ref{eq:nonlin}) where  $\odot$ is the element-wise product of the incoming covariance matrix, $\boldsymbol{\Sigma}_z$, and the squared gradient, $\nabla$, of non-linear function with respect to the incoming mean, $\boldsymbol{\mu}_z$. 
\begin{equation}\label{eq:nonlin}
    \begin{split}
    \boldsymbol{\mu}_g &\approx \Psi(\boldsymbol{\mu}_z) \\
    \boldsymbol{\Sigma}_g &\approx \boldsymbol{\Sigma}_z \odot \nabla\Psi(\boldsymbol{\mu}_z)\nabla\Psi(\boldsymbol{\mu}_z)^T
    \end{split}
\end{equation}
For the Softmax classification layer at the output of the network, let $\boldsymbol{\mu_{\hat{y}}}$ and $\boldsymbol{\sigma^2_{\hat{y}}}$ denote the mean and variance, respectively, that will be used to infer the ELBO objective function.
\subsection{Closed Form ELBO}
Using the propagated values $\boldsymbol{\mu_{\hat{y}}}$ and $\boldsymbol{\sigma^2_{\hat{y}}}$, the ELBO, Equation (\ref{eq:ELBO_der}), can be written in closed form, as shown in Equation (\ref{eq:ELBO_closed}). The weighting variable $\boldsymbol{\tau}$ is added to control the level of explicit regularization toward the prior induced by the KL divergence term. The KL divergence is computed on a per-parameter basis and then summed, where $\boldsymbol{|\Omega|}$ denotes the cardinality of the set $\boldsymbol{\Omega}$, i.e., the number of weight parameters, and $N$ denotes the number of classes or output nodes of the final layer.

\begin{figure*}[b]
\centerline{\includegraphics[width=0.9\textwidth]{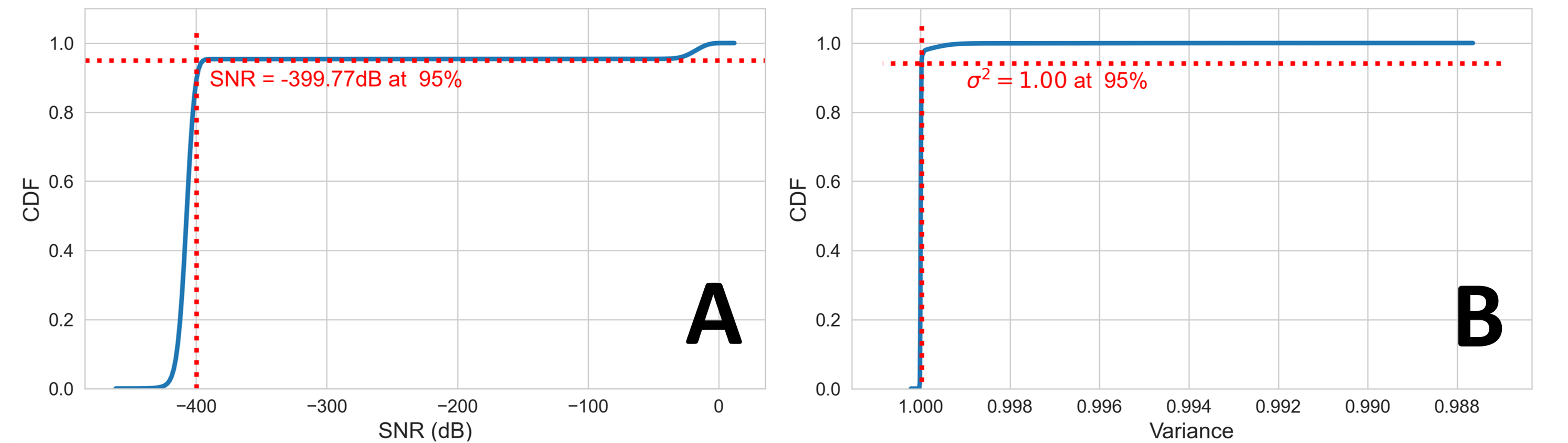}}
\caption{Analysis of parameter uncertainty from a two 800-node hidden layer fully connected network (A) Cumulative Distribution Function plot of the Signal-to-Noise Ratio (SNR) demonstrating $95\%$ of the parameters are approximately -400dB SNR. (B) Cumulative Distribution Function plot of the Variance demonstrating $95\%$ of the parameters have a variance of 1 or greater.}
\label{fig:CDFs}
\end{figure*}

\subsection{Training Moment Propagation}
Contrary to other traditional probabilistic problems where VI can be applied, estimating a prior for a DNN is difficult before any training has occurred. Given the dual objective of the ELBO, reverting to a prior far from the parameterization of the true posterior would be detrimental to the learning process. Instead, sparsity-inducing priors, such as the standard normal, $p(\Omega_i) = \mathcal{N}\left(\mu_{p_i}=0,\sigma_{p_i}^2=1\right)$, can be used to remove parameters that are not required to maximize the model likelihood by reverting their mean and variance, to that of the sparsity inducing prior. 

Thus, the maximum variance of a parameter present in the network will be $\sigma_{\omega_i}^2 = 1$. The incorporation of a sparsity-inducing prior in the ELBO objective aligns with the Minimum Description Length principle. This is achieved by inherently reducing the complexity of a DNN, specifically by systematically eliminating parameters that are not necessary for a given task \cite{hinton_keeping_1993, He2020}. 
\begin{strip}
    \fontsize{9.5}{8}\selectfont
    \begin{equation}\label{eq:ELBO_closed}
    \text{ELBO} = -\frac{N}{2}\ln{(2\pi)} - \frac{1}{2}\ln{(|\boldsymbol{\Sigma_{\hat{y}}}|)} - \frac{1}{2}\left({(\boldsymbol{y} - \boldsymbol{\mu_{\hat{y}}})}^T\boldsymbol{\Sigma_{\hat{y}}}^{-1}(\boldsymbol{y} - \boldsymbol{\mu_{\hat{y}}})\right) 
    - \frac{\tau}{2}\sum^{\boldsymbol{|\Omega|}}_{i=1}{\left(-1 + \frac{(\mu_{q_{\omega_i}} - \mu_{p_i})^2}{\sigma^2_{q_{\omega_i}}} + \ln\left(\frac{\sigma_{p_i}^2}{\sigma_{q_{\omega_i}}^2}\right) + \frac{\sigma^2_{q_{\omega_i}}}{\sigma^2_{p_i}}\right)}
\end{equation}
\end{strip}

\section{Weight Uncertainty}
After training, parameters essential for a task are expected to exhibit reduced uncertainty. Conversely, parameters unessential for the task are expected to gravitate towards the sparsity-inducing prior, resulting in increased uncertainty. To measure a parameter's relative importance to a DNN's function, two forms of uncertainty are considered: the variance of each parameter, $\sigma_{\omega_i}^2$, and the Signal-to-Noise Ratio (SNR) of each parameter, $\text{SNR}_{\omega_i}=|\mu_{\omega_i}|/\sigma_{\omega_i}^2$. Parameter importance is inversely proportional to a parameter's variance and proportional to a parameter's SNR. MP's ability to self-determine important parameters is demonstrated by observing a cumulative distribution function (CDF) of both parameter SNR and variance from a trained two 800-node hidden layer fully connected network. The CDFs for parameter uncertainty are then correlated to the network's ability to prune parameters based on each parameter's learned uncertainty. 

The CDFs for the uncertainty measurements are presented in Figure \ref{fig:CDFs}. Parameter importance based on SNR and Variance demonstrates that $95\%$ of the fully connected network parameters are uncertain and, thus, are considered unimportant. Similarly, the CDF of the variance of the parameters demonstrates that $95\%$ of the parameters have been optimized to be equivalent to the prior and are thus uncertain. 

Although parameters aligning with the prior may appear equivalent, it doesn't necessarily imply insignificance to the network's functional approximation. To demonstrate parameter importance to a learned network function, parameters are ordered and pruned according to their SNR and variance at various percentages of the total network parameters. The subsequent impact on network performance is then assessed.

During pruning, all network parameters are ordered based on importance, regardless of layer, as there is no guarantee that important parameters will be evenly distributed across all layers. This approach allows for more efficient pruning without compromising performance. The performance of the pruned network, considering SNR and variance-based pruning, is compared against random pruning (the lower performance bound) and pruning based on the smallest absolute value. Additionally, the performance of the MP framework is compared to Bayes-by-Backprop (BBB) and standard deterministic networks, both featuring the same two hidden layer architecture. For MP and BBB frameworks, smallest absolute value-based pruning is performed on parameters using the mean of the parameter's approximating distribution. 

\begin{figure*}[t]
\centerline{\includegraphics[width=0.6\textwidth]{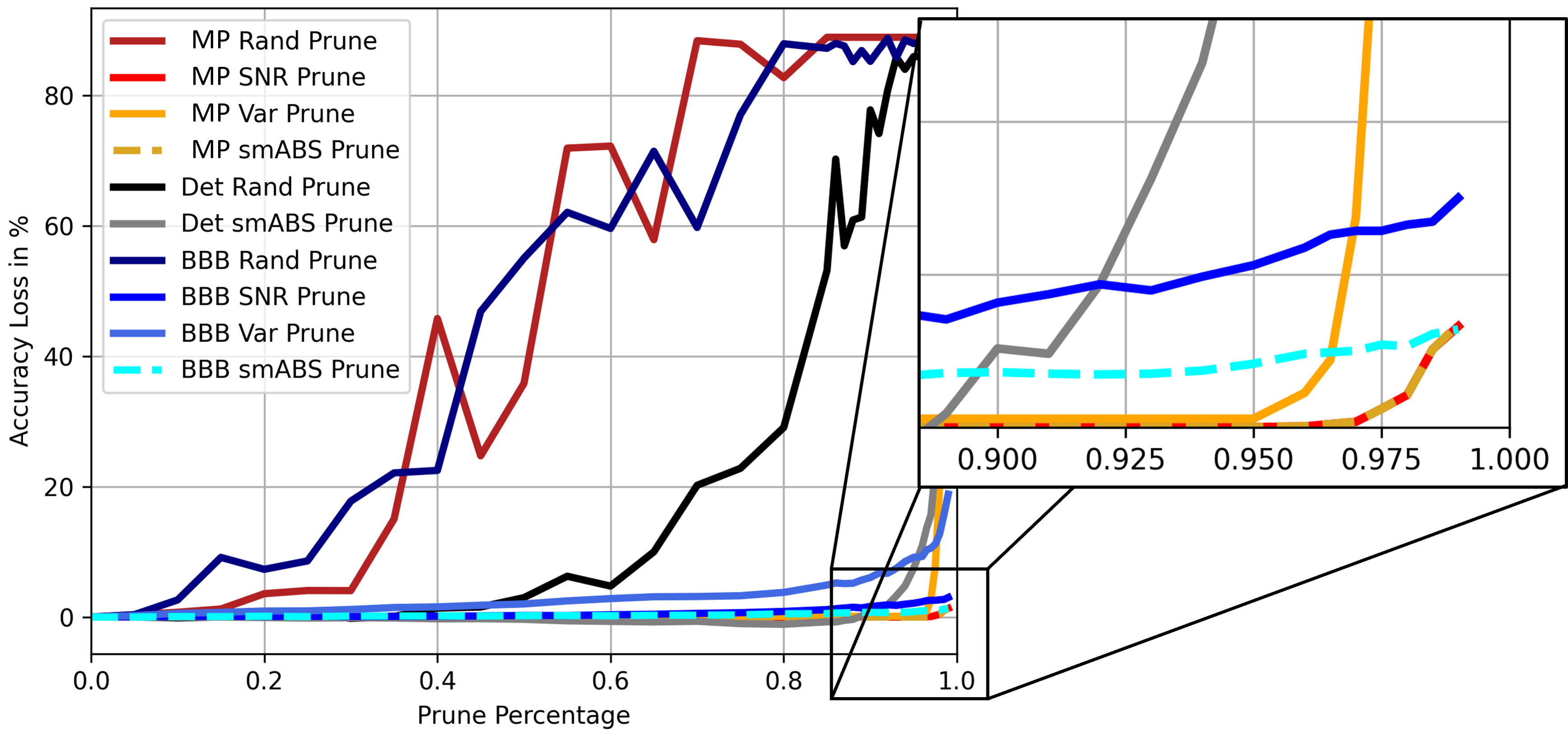}}
\caption{Loss in performance from the original validation accuracy as a result of various pruning methods. Moment Propagation performance is presented in warm colors, Bayes-by-Backprop (BBB) performance in cool colors, and deterministic performance in grey and black.}
\label{fig:pruning}
\end{figure*}

Figure \ref{fig:pruning} shows that the MP network maintained more performance, even with a higher percentage of the network pruned compared to the BBB and deterministic frameworks. Although variance-based pruning exhibited a greater overall loss in performance compared to SNR-based pruning for MP, both approaches began to experience performance degradation at 95\% of the network being pruned, as reflected in the CDFs for variance and SNR. Additionally, in the BBB network, variance-based pruning performed significantly worse than SNR-based pruning, indicating that variance may not be as reliable as a parameter importance measure in the BBB framework. 

Based on the pruning performance of the MP framework, trained MP networks can accurately discern important parameters via learned parameter uncertainty. This inherent ability for self-determination of relevant parameters is a crucial tool for CL. 
Important parameters can be appropriately regularized to avoid changes and prevent catastrophic interference, ensuring the retention of knowledge from previous tasks. In contrast,  unimportant parameters, which have demonstrated minimal contribution to network predictions, can be used for learning subsequent information, making MP beneficial in Continual Learning scenarios.

\section{Bayesian Continual Learning}
Continual Learning (CL) can be divided into three sub-categories: \textit{Task Incremental Learning},\textit{ Domain Incremental Learning}, and \textit{Class Incremental Learning}, where the premise of learning and retaining information over multiple training periods remains the same, but how context is provided changes in each scenario \cite{van_de_ven_three_2022}. Methods described in this paper will focus on \textit{Task Incremental Learning}, in which the context $\boldsymbol{\mathscr{C}}$ for the input $\boldsymbol{\mathscr{X}}$ changes over time and is provided during training and inference periods. The output space $\boldsymbol{y}$ is separated for each input context, producing the mapping $\boldsymbol{f}: \boldsymbol{\mathscr{X}} \times \boldsymbol{\mathscr{C}} \rightarrow \boldsymbol{y}$ \cite{van_de_ven_three_2022}. This additional task context is only leveraged for the multi-headed network architecture in which task information is required to choose the appropriate output head and is not used to select which network parameters are active.

\subsection{Learning Rate Adaptation}
Initially explored by Ebrahimi et al.\cite{ebrahimi_uncertainty-guided_2020}, \textit{Learning Rate Adaptation} (LRA) leverages learned parameter uncertainty to adapt the learning rates of individual parameters according to their relevance to the network's functional approximation. Lower learning rates for important network parameters ideally prevents catastrophic interference in these parameters by restricting change while allowing unimportant parameters to learn new information freely. Our approach to Learning Rate Adaption replaces the Bayes-by-Backprop (BBB) framework with the Moment Propagation framework. Instead of using a BBB parameter's latent sampling distribution to determine parameter importance, parameter importance is determined by a random variable that directly contributes to the network's predictive distribution.

Learning rates for each parameter are determined by mapping parameter importance from all network parameters to a user-defined range of learning rates. First, parameter importance is determined according to the parameter's variance, $1/\sigma_{\omega_i}^2$, or SNR, $|\mu_{\omega_i}|/\sigma_{\omega_i}^2$. The resulting importance values are remapped across the whole network, excluding the classification head, instead of across each layer given there is no guarantee that parameter importance will be even distributed across all layers. The remapping function is shown in Equation (\ref{eq:LRImport}), where $\boldsymbol{\alpha}_{t+1}$ represents the vector of all parameter learning rates for the next task, $\boldsymbol{\iota}_t$ represents the vector of all parameter importance values from task, t. The min and max user-defined range of learning rate values are defined as $\alpha_{\text{min}}$ and $\alpha_{\text{min}}$. Learning rates for each parameter's mean and variance are updated synchronously because the approximating distribution for each parameter is treated as a single parameter. Ultimately, the mean and variance of the most important parameters will receive the lowest user-defined learning rate, restricting any change in the parameter. The mean and variance of the most uncertain parameters will receive the highest, allowing these parameters to learn freely. 
\begin{equation}
\label{eq:LRImport}
    \boldsymbol{\alpha}_{t+1} \gets \frac{((\boldsymbol{\iota}_t - \text{min}(\boldsymbol{\iota}_t))(\alpha_{\text{max}} - \alpha_{\text{min}})}{\text{max}(\boldsymbol{\iota}_t) - \text{min}(\boldsymbol{\iota}_t)} + \alpha_{\text{min}}
\end{equation}

\subsection{Per-Parameter Bayesian Inference}
Our second approach, \textit{Per-Parameter Bayesian Inference} (PPBI), leverages the same concept of leveraging parameter uncertainty but instead changes the weighting of the KL regularization term within the ELBO. Drawing inspiration from Ebrahimi et al. \cite{ebrahimi_uncertainty-guided_2020}, Ahn et al. \cite{ahn_uncertainty-based_2019}, and Nguyen et al. \cite{nguyen_variational_2018}, this framework performs approximate Bayesian Inference on a per-parameter basis guided by parameter uncertainty without any changes to the regularization term. PPBI applies a similar methodology to Learning Rate Adaption; however, for each new task, a KL regularization term weighting value is applied according to each parameter's importance. Before training a task the network prior is replaced with the previous tasks learned posterior. To control the explicit regularization of every parameter, regularization weights are adjusted based on each parameter's importance by mapping parameter importance values to a user-defined range of regularization weighting values. For this technique, the most important parameters will map to the highest user-defined weighting, heavily restricting change from the previous tasks posterior (the prior), while the least important parameters will map to the lowest user-defined weighting, allowing those parameters to change easily to maximize the network log-likelihood for the current task. For PPBI, the remapping function is shown in Equation (\ref{eq:KLImport}).

\begin{equation}
\label{eq:KLImport}
    \boldsymbol{\tau}_{t+1} \gets \frac{((\boldsymbol{\iota}_t - \text{min}(\boldsymbol{\iota}_t))(\tau_{\text{min}} - \tau_{\text{max}})}{\text{max}(\boldsymbol{\iota}_t) - \text{min}(\boldsymbol{\iota}_t)} + \tau_{\text{max}}
\end{equation}

The algorithm for MP-based CL for both LRA and PPBI is presented in Algorithm \ref{alg:CL}. 

\begin{algorithm}
\caption{Moment Propagation Continual Learning}\label{alg:CL}
\begin{algorithmic}[1]
\Require $\{\mathcal{D}_t\}_{t\in\boldsymbol{\mathscr{T}}}; \quad \mathcal{D}_t=(\boldsymbol{\mathscr{X}}_t,\boldsymbol{y}_t)$ \par 
\hskip\algorithmicindent $q_{\boldsymbol{\theta}_0}(\boldsymbol{\Omega_0}) \sim \prod^{|\boldsymbol{\Omega}_0|}_{j=1}{\mathcal{N}(\mu_{w_0^{(j)}},\sigma_{w_0^{(j)}}^2)}$\par
\hskip\algorithmicindent$p_0(\boldsymbol{\Omega}_0) \sim  \prod^{|\boldsymbol{\Omega}_0|}_{j=1}{\mathcal{N}(0,1)}$ \par
\hskip\algorithmicindent $\boldsymbol{\tau}_0 = (\tau_0^{(j)} \cdots \tau_0^{(j)})$ \par 
\hskip\algorithmicindent$\boldsymbol{\alpha}_0 = (\alpha_0^{(j)} \cdots \alpha_0^{(j)})$ 
\For{all tasks, $t$,} 
    \Repeat
        \For{all examples in $\mathcal{D}_t$} 
        \State $\mathcal{L}_{\text{LL}} \gets \mathop{\mathbb{E}_{q_{\boldsymbol{\theta}_t}(\boldsymbol{\Omega}_{t})}}[ \ln{p(\boldsymbol{\hat{y}}|\boldsymbol{\mathscr{X}}_t,\boldsymbol{\Omega}_{t})}]$ 
        \State $\mathcal{L}_{\text{KL}} \gets \sum_{j=1}^{|\boldsymbol{\Omega}_t|} {\tau^{(j)}_t\text{KL}_{q_{\boldsymbol{\theta}_t}(\boldsymbol{\Omega}^{(j)}_{t})}[q_{\boldsymbol{\theta}_t}(\boldsymbol{\Omega}^{(j)}_{t})||p_t(\boldsymbol{\Omega}^{(j)}_t)]}$ 
        \State $\mathcal{L}_{\text{VDP}} \gets \mathcal{L}_{\text{LL}} - \mathcal{L}_{\text{KL}}$ 
        \For{each network parameter, $\boldsymbol{\Omega}_t^{(j)}$,}
            \State $\boldsymbol{\Omega}_t^{(j)} \gets \boldsymbol{\Omega}_t^{(j)} - \boldsymbol{\alpha}^{(j)}_t\frac{\partial\mathcal{L}_{\text{VDP}}}{\partial\boldsymbol{\Omega}_t^{(j)}}$ 
        \EndFor
        \EndFor
    \Until{validation accuracy plateaus} 
    \For{each network parameter, $\boldsymbol{\Omega}_t^{(j)}$,}
        \State $\iota^{(j)}_t \gets 1/\sigma_{w_j}^2$ or $\mu_{w_j}/\sigma_{w_j}^2$
    \EndFor
    \If{Learning Rate Adaptation}
        \State $\boldsymbol{\alpha}_{t+1} \gets \frac{((\boldsymbol{\iota}_t - \text{min}(\boldsymbol{\iota}_t))(\alpha_{\text{max}} - \alpha_{\text{min}})}{\text{max}(\boldsymbol{\iota}_t) - \text{min}(\boldsymbol{\iota}_t)} + \alpha_{\text{min}}$
        \State $\boldsymbol{\tau}_{t+1} \gets \boldsymbol{\tau}_{t}$
    \EndIf 
    \If{Per-Parameter Bayesian Inference}
        \State $\boldsymbol{\tau}_{t+1} \gets \frac{((\boldsymbol{\iota}_t - \text{min}(\boldsymbol{\iota}_t))(\tau_{\text{min}} - \tau_{\text{max}})}{\text{max}(\boldsymbol{\iota}_t) - \text{min}(\boldsymbol{\iota}_t)} + \tau_{\text{max}}$
        \State $p_{t+1}(\boldsymbol{\Omega}_{t+1}) \gets q_{\boldsymbol{\theta}_t}(\boldsymbol{\Omega}_{t})$
        \State $\boldsymbol{\alpha}_{t+1} \gets \boldsymbol{\alpha}_{t}$
    \EndIf 
\EndFor
\end{algorithmic}
\end{algorithm}

\subsection{Experimental Setup}
\subsubsection{Datasets and Networks}
Our CL methodologies are evaluated for eight different CL benchmark datasets of increasing difficulty: Two Split MNIST, Five Split MNIST, Permuted MNIST, Two Split CIFAR10, Five Split CIFAR10, Mixed CIFAR10-CIFAR100, and a sequence of eight datasets. Split MNIST and CIFAR10 datasets consist of each base dataset separated into two tasks of five classes and five tasks of 2 classes, for two split and five split, respectively. The Permuted MNIST dataset consists of ten different pixel level permutations applied to the entire base MNIST dataset, resulting in ten tasks of ten classes each. Mixed CIFAR10-CIFAR100 combines the base CIFAR10 and CIFAR100 benchmark datasets and alternates between tasks of two classes from CIFAR10 and twenty classes from CIFAR100, for ten tasks. Finally, a sequence of eight datasets is evaluated, consisting of MNIST, CIFAR10, CIFAR100, NotMNIST, SVHN, Traffic Signs, FaceScrub, and FashionMNIST where each task is a new dataset. 

For all sequential approaches, 15\% of the training set for each task is reserved for validation, while the test set is exclusively used for testing after task training is complete. The order of each dataset is randomized for each epoch, including for joint training. All datasets are normalized to the mean and the standard deviation of the full dataset before splitting. For split CIFAR10 and mixed CIFAR10/CIFAR100, a random crop with a padding of 4 pixels and a random horizontal flip are added to assist with generalization. These transforms are added to existing approaches for a fair comparison. Transform was not applied to the sequence of eight datasets, as it adversely affected performance, but datasets are padded to a image size of 32x32 pixel, and input data consisting of one channel are replicated across two additional channels for a consistent input image size of 32x32x3. 

A two 800-node hidden layer fully connected network architecture is used across all approaches to compare continual learning performance for two-split, five-split, and permuted MNIST datasets. An AlexNet Convolutional Neural Network architecture is used for two-split and five-split CIFAR10, mixed CIFAR10-CIFAR100, and the sequence of eight datasets. We recollect all results for these architectures and datasets, slightly improving on some previously reported results due to differences in architectures used. 

\subsubsection{Hyperparameters}
A grid search is performed for hyperparameters to maximize the performance across all tasks. Given that the variance of each parameter directly contributes to the predictive distribution, the initialization of these values has a significant impact on overall performance. Thus, parameter variance initialization is searched between $\sigma^2_\pi \gets [-10, -18]$.

\begin{table*}[b]
\setlength\tabcolsep{3pt} 
\caption{Task Incremental Learning Results}
\begin{center}
\begin{tabular}{|c|cc|cc|cc|cc|cc|cc|cc|}
  \hline
  \multirow{2}{*}{\textbf{Approach}} & \multicolumn{2}{c}{\textbf{2-Split MNIST}} & \multicolumn{2}{|c}{\textbf{5-Split MNIST}} & \multicolumn{2}{|c}{\textbf{Permuted MNIST}} & \multicolumn{2}{|c}{\textbf{2-Split CIFAR10}} & \multicolumn{2}{|c|}{\textbf{5-Split CIFAR10}} & \multicolumn{2}{|c|}{\textbf{CIFAR10/100}} & \multicolumn{2}{|c|}{\textbf{Sequence}} \\
  \cline{2-15}
  & \textbf{ACC} & \textbf{BWT} & \textbf{ACC} & \textbf{BWT} & \textbf{ACC} & \textbf{BWT} & \textbf{ACC} & \textbf{BWT} & \textbf{ACC} & \textbf{BWT} & \textbf{ACC} & \textbf{BWT} & \textbf{ACC} & \textbf{BWT} \\
  \hline
  EWC       & 98.12\% & -1.03\% & 99.18\% & -0.57\% & 96.19\% & -0.46\% & 82.78\% &  -0.48\% & 86.01\% & -3.17\% & 69.28\% & -4.20\% & 70.81\% & -3.05\% \\
  SI        & 98.85\% & -0.08\% & 99.47\% &  -0.24\% &  93.88\% &  -0.06\% &  86.19\% &   -1.49\% &  86.85\% &  -5.74\% & 70.39\% & -9.04\% & 61.91\% & -16.04\% \\
  MAS       & 98.82\% & -0.13\% & 99.57\% &  -0.01\% &  95.46\% &  -0.21\% &  99.46\% &   -2.28\% &  85.98\% &  -7.00\% & 70.70\% & -8.56\% & 60.62\% & -13.84\% \\
  UCL       & 98.66\% & -0.41\% & 99.21\% & -0.56\% & 94.97\% & -2.58\% & 82.02\% &  -2.46\% & 84.26\% & -6.78\% & 71.68\% & -4.16\% & 69.81\% & -2.10\% \\
  UCB       & 99.18\% & \textbf{+0.01\%} & 99.63\% &  0.00\% & 97.42\% &  \textbf{0.00\%}& 73.76\% &  -5.10\% & 76.67\% & -7.56\% & 60.64\% & -15.02\% & 48.37\% & -24.90\% \\
  LRA (Var.)    & 99.32\% & -0.04\% & 99.85\% & \textbf{+0.02\%} & 97.40\% & -0.81\% & \textbf{90.83\%} &  \textbf{-0.30\%} & \textbf{92.17\%} & -0.28\% & \textbf{77.16\%} & -1.43\% & \textbf{77.40\%} & -0.64\% \\
  LRA (SNR)    & 99.29\% & -0.02\% & 99.80\% & -0.03\% & 97.65\% & -0.44\% & 89.61\% &  -0.96\% & 91.69\% & -0.91\% & 76.58\% & -4.22\% & 76.94\% & -1.15\% \\
  PPBI (Var.)    & \textbf{99.40\%} & -0.05\% & 99.81\% & -0.06\% & \textbf{98.19\%} & -0.25\% & 89.29\% &  -0.60\% & 89.78\% & \textbf{+0.04\%} & 75.12\% & -1.31\% & 77.04\% & \textbf{-0.10\%} \\
  PPBI (SNR)    & 99.37\% & -0.06\% & \textbf{99.84\%} & -0.04\% & 98.18\% & -0.27\% & 89.46\% &  -1.02\% & 89.99\% & -0.06\% & 75.02\% & \textbf{-0.29\%} & 77.15\% & -0.61\% \\
  \Xhline{4\arrayrulewidth}
   HAT       & 98.79\% &  0.00\% & 99.75\% &  0.00\% & 97.34\% &  0.00\% & 86.91\% &   0.01\% & 92.09\% &  0.00\% & 77.84\% & -0.03\% & 83.13\% & -0.03\% \\
  \Xhline{4\arrayrulewidth}
  MP-FT & 98.98\% & -0.20\% & 98.32\% & -1.46\% & 95.91\% & -2.47\% & 85.72\% & -6.30\% & 79.41\% & -15.46\% & 57.80\% & -19.55\% & 62.98\% & -20.44\% \\
  MP-FF & 98.91\% &  0.00\% & 99.40\% &  0.00\% & 96.72\% &  0.00\% & 88.72\% &  0.00\% & 89.54\% &   0.00\% & 69.27\% &  -0.25\% & 76.70\% & -0.01\% \\
  MP-JT & 99.39\% &  0.00\% & 99.73\% & -0.08\% & 98.15\% & -0.17\% & 91.51\% & +1.14\% & 94.64\% &  +0.79\% & 87.46\% &  +1.23\% & 78.27\% & -1.08\% \\
  \hline
  \end{tabular}
\label{tab:1}
\end{center}
\end{table*}

Regularization toward the prior also has a significant impact on performance. The initial KL weighting is searched between $\tau_0 \gets [1e\text{-}3, 1e\text{-}8]$ and is initially applied to all parameters regardless of approach. Values higher than $1e\text{-}3$ typically cause too much regularization toward the prior, resulting in poor performance due to the network's inability to learn a sufficiently complex representation for the task. For LRA, the maximum learning rate for the mapping to the user-defined range is searched between $\alpha_{\text{max}} \gets [1e\text{-}3, 1e\text{-}5]$. For KL Weight Adaptation, the maximum KL weighting for the mapping to the user-defined range is searched between $\tau_{\text{max}} \gets [1e\text{-}2, 1e\text{-}7]$. The minimum learning rate and KL weighting are both set to $1e\text{-}12$ and are not tuned. The number of epochs and batch size were fixed at 250 and 500, respectively. A large batch size was chosen because more performance was retained in deterministic baseline tests with Fine Tuning and enabled more efficient use of computational resources. 
\subsection{Performance Measurement}
Performance is gauged through the Average Test Classification Accuracy (ACC) and Backward Transfer (BWT). Average Test Classification Accuracy is an average of all test accuracies on individual tasks after training all tasks. Backward Transfer indicates how much learning new information has affected performance on previous tasks. Backward Transfer values less than zero indicate catastrophic forgetting, while values greater than zero indicate improved performance on previous tasks after training on new information \cite{ebrahimi_uncertainty-guided_2020}. These metrics are shown in Equation \ref{eq:metrics}.

\begin{equation}\label{eq:metrics}
    \begin{split}
        \text{BWT} &= \frac{1}{t}\sum^{t}_{i=1}R_{i,t} - R_{i,i} \\
        \text{ACC} &= \frac{1}{t}\sum^{t}_{i=1}R_{i,t}
    \end{split}\tag{14}
\end{equation}

\section{Results and Discussion}
Our CL methodologies are compared to parameter uncertainty-based methods: Uncertainty-Based Continual Learning (UCL)\cite{ahn_uncertainty-based_2019} and Uncertainty-Guided Continual Learning (UCB)\cite{ebrahimi_uncertainty-guided_2020}. We also compare our methods to long-standing standards for continual learning: Elastic Weight Consolidation (EWC) \cite{kirkpatrick_overcoming_2017}, Memory Aware Synapses (MAS)\cite{aljundi_memory_2018}, Synaptic Intelligence (SI)\cite{zenke_continual_2017}, and Hard Attention to Task (HAT)\cite{serra_overcoming_2018}. These approaches to catastrophic interference mitigation are compared to baselines using Moment Propagation Framework: Fine Tuning (FT), where no efforts are made to mitigate catastrophic interference, Feature Freezing (FF), where all but the classification head is frozen after training the first task, and Joint Training (JT), where tasks are trained sequentially, but jointly with previous tasks. Fine Tuning represents the lower bound on performance, indicating performance was not gained over sequential training, while Joint Training represents the theoretical upper bound on performance. The performance results for all listed methods are presented in Table \ref{tab:1}, where the maximum ACC and maximum BWT values for each dataset are shown in bold. 

Our MP-based LRA and PPBI methodologies outperform their sampling-based predecessors, UCB and UCL, respectively. This performance improvement can be attributed to better measures of parameter uncertainty resulting from the MP framework. In UCB, performance improved as the number of samples of the predictive distribution increases \cite{ebrahimi_uncertainty-guided_2020}. Approximating the predictive distribution via the propagated moments improves measures of the network log-likelihood. The directly differentiable network log-likelihood then improves gradient updates of the network parameter, improving the measure of parameter uncertainty and, thus, parameter importance. Similarly, UCL only leverages one sample of the predictive distribution to estimate the network likelihood \cite{ahn_uncertainty-based_2019} but outperforms UCB on more complex benchmark datasets by heavily relying on changes made to the KL regularization term in the ELBO to further restrict parameter change. Additionally, our methods perform on par with HAT's Average Test Accuracy despite not having a propagation mask to select which features are used for which tasks. This masks helps HAT maintain a BWT of near zero by not only freeze important parameters to restrict parameter change, but also masking out all other unimportant parameters to prevent interference on a feature level. 

Despite improved performance over previous approaches, results for LRA and PPBI only show a marginal performance improvement over hyperparameter tuning the FF method. This implies that freezing a MP DNN after training the first task and learning a new classification layer can provide reasonable Task Incremental Learning performance. This trend differed slightly for the mixed CIFAR10 CIFAR100 dataset, where the ACC for FF was lower than uncertainty-based methods, indicating the network did not gain enough information from the training of the first task, two classes of CIFAR10, to sufficiently adapt to CIFAR100 based tasks with only learning the classification layer. 

\subsection{Split and Permuted MNIST}
PPBI slightly outperforms LRA for split and permuted MNIST benchmark sets. For two-split MNIST, PPBI achieves an Average Test Accuracy of $99.40\%$ and a Backward Transfer of $-0.05\%$ with a Variance-based uncertainty metric. For five-split MNIST, PPBI achieves an Average Test Accuracy of $99.84\%$ and a Backward Transfer of $-0.04\%$ with an SNR-based uncertainty metric. For permuted MNIST with ten tasks, PPBI achieves an Average Test Accuracy of $98.19\%$ and a Backward Transfer of $-0.25\%$ with a variance-based uncertainty metric. The restrictive nature of Bayesian Inference minimizing the KL divergence between the variational posterior and prior may have helped in this scenario, given that all tasks are distributed. Thus, restricting change from one task to the next may have had less impact than more advanced tasks. 
\subsection{Split CIFAR10 and Mixed CIFAR10/100}
Conversely, for Split CIFAR10 and Mixed CIFAR10/CIFAR100, LRA sightly outperforms PPBI. For two-split CIFAR10, LRA achieves an Average Test Accuracy of $90.83\%$ and a Backward Transfer of $-0.30\%$ with a Variance-based uncertainty metric. For five-split CIFAR10, LRA achieves an Average Test Accuracy of $92.17\%$ and a Backward Transfer of $-0.28\%$ with a Variance-based uncertainty metric. For mixed CIFAR10 CIFAR100, LRA achieves an Average Test Accuracy of $77.16\%$ and a Backward Transfer of $-0.29\%$ with a Variance-based uncertainty metric. We attribute these results to the same characteristics of Per-parameter Bayesian Inference, which benefited Split and Permuted MNIST tasks. However, in the case of CIFAR10 and CIFAR100 datasets separating classes into tasks may require parameters to diverge from the previous tasks posterior. Thus, PPBI may cause unwanted regularization to the network prior. Learning Rate Adaptation avoids this problem by restricting all changes in important parameters while allowing unimportant parameters to move away from the network posterior to the previous task.

\subsection{Sequence of Eight Datasets}
For the Sequence of Eight Datasets PPBI and LRA, perform similarly. LRA provides the maximum Average Test Accuracy at $77.40\%$ with a Backward Transfer of $-0.64\%$ with a variance-based uncertainty metric. PPBI and LRA significantly outperform all other methods apart from HAT.

\section{Conclusion}
In this work, two Continual Learning methodologies are presented that leverage learned parameter uncertainty derived from a Moment Propagation framework to regularize training of new tasks to prevent Catastrophic Forgetting. These methods are detailed by deriving the custom layers for a basic Convolutional Neural Network from the principles of Variational Inference. The concept of parameter importance through learned uncertainty from the Moment Propagation framework's is demonstrated and applied to Continual Learning through two methodologies to Learning Rate Adaptation and Per-parameter Bayesian Inference. While these approaches leverage learned parameter importance in different ways, both mitigate catastrophic forgetting through regulariz the learning of network parameters. These methods were evaluated on multiple sequential benchmark datasets, and performance was compared to other similar previously published approaches. Ultimately, Learning Rate Adaptation and Per-parameter Bayesian Inference outperform or yield comparable results to existing approaches through improved measures of parameter uncertainty.

\bibliographystyle{ieeetr}
\bibliography{refs}

\end{document}